\documentclass{article}
\usepackage{ijcai20}

\usepackage{times}
\usepackage{soul}
\usepackage{url}
\usepackage[hidelinks]{hyperref}
\usepackage[utf8]{inputenc}
\usepackage[small]{caption}
\usepackage{graphicx}
\usepackage{amsmath}
\usepackage{amsthm}
\usepackage{booktabs}
\usepackage{bookmark}
\usepackage{algorithm}
\usepackage{algorithmic}
\algsetup{linenosize=\scriptsize}

\usepackage{amsmath}
\usepackage{subcaption}
\usepackage{amsfonts}
\usepackage{amssymb}
\usepackage{bm}
\usepackage{mathtools}
\usepackage{amsthm}
\usepackage{multirow}
\usepackage{tabularx}
\usepackage{url}
\usepackage{xcolor}
\usepackage{enumitem}

\definecolor{pastelgreen}{rgb}{0.01, 0.75, 0.24}
\definecolor{red(pigment)}{rgb}{0.93, 0.11, 0.14}
\definecolor{bleudefrance}{rgb}{0.19, 0.55, 0.91}

\newcommand{\mysection}[1]{\vspace{1pt}\noindent\textbf{#1}\hspace{10pt}}

\newtheorem{dfn}{Definition}

\newtheorem{prop}{Proposition}
\theoremstyle{plain}

\newcommand{\citet}[1]{\citeauthor{#1}~\shortcite{#1}}

\title{Signaling Friends and Head-Faking Enemies Simultaneously: \\ Balancing Goal Obfuscation and Goal Legibility}

\author{
Anagha Kulkarni 
\And
Siddharth Srivastava 
\And
Subbarao Kambhampati 
\affiliations
CIDSE, Arizona State University, Tempe, AZ 85281 USA 
\emails
\{anaghak, siddharths, rao\}@asu.edu
}

\begin{document}

\maketitle

\begin{abstract}
\vspace{-6pt}
In order to be useful in the real world, an AI agent needs to plan and act in the presence of other agents, who may be helpful or disruptive. In this paper, we consider the problem where an autonomous agent needs to act in a manner that clarifies its objectives to cooperative agents while preventing adversarial agents from inferring those objectives. This problem is solvable when the agent has access to the sensor models used by the cooperative and adversarial agents and access to their prior knowledge. We develop two new solution approaches: one provides an optimal solution to the problem given a fixed time horizon by using an integer programming solver, the other provides a satisficing solution using heuristic-guided forward search to achieve prespecified amount of obfuscation and legibility for adversarial and cooperative agents respectively. We show the feasibility and utility of our algorithms through extensive empirical evaluation on planning benchmarks.
\end{abstract}

\vspace{-17pt}

\section{Introduction} 

In a multi-agent environment, the activities performed by an agent may be observed by other agents. In such an environment, an agent should perform its tasks while taking into account the observers' sensing capabilities and its relationship with the observers. Several prior works have explored the generation of legible agent behavior in the presence of cooperative observers \cite{Dragan-RSS-13,exp-yz,macnally2018action} and obfuscating agent behavior in the presence of adversarial observers \cite{keren2016privacy,ijcai2017-610,shekhar2018representing,implicitHRC2018}. Through legible behavior, an agent can convey the necessary information to a cooperative observer, while through obfuscating behavior it can hide sensitive information from an adversary. However, these prior works assume that the observers in the environment are either entirely cooperative or entirely adversarial. 

In real-world settings of strategic importance, an agent might encounter both types of observers \emph{simultaneously}. This would necessitate synthesizing a behavior that is simultaneously legible to friendly entities and obfuscatory to adversarial ones. For instance, in soccer, a player may perform feinting trick to confuse an opponent while signaling a teammate. 
Synthesizing a single behavior that is legible and obfuscatory to different agents presents significant technical challenges.
In particular, the agent may have to be deliberately less legible to its friends so that it can be effectively more obfuscatory to its adversaries. 
This problem gives rise to a novel optimization space that involves trading-off the amount of obfuscation desired for adversaries with the amount of legibility desired for friends. 


In this paper, we present a novel problem framework called \emph{mixed-observer controlled observability planning problem}, \textsc{mo-copp}, that allows an agent to simultaneously control information yielded to both cooperative and adversarial observers while achieving its goal. 
Our framework models and exploits situations where different observers have differing sensing capabilities, which result in different ``perceptions" of the same information.
Typically, different observers in an environment may have different ``sensors" (perception capabilities) due to differences in prior communication, background knowledge, and raw sensor hardware. For example, an agent may establish semaphore actions that are more meaningful to its allies than to its adversaries. Therefore, we assume that different types of observers have differing sensor models.

We propose two approaches to solve \textsc{mo-copp}. \textbf{(1)} A novel integer programming (IP) encoding, which provides an optimal solution given a fixed time horizon. This involves maximizing the amount of trade-off between obfuscation achieved for adversarial observer and legibility achieved for cooperative observer with respect to the agent's objectives. \textbf{(2)} A heuristic-guided search algorithm, which leverages a prior approach \cite{implicitHRC2018} that was designed to address entirely cooperative or entirely adversarial settings. Through theoretical and empirical analysis, we explore the strengths of the two proposed approaches. Additionally, we show that for mixed-observer setting, our solution approaches add significant value over approaches that consider either entirely cooperative or adversarial observers. 

\begin{figure}[!t]
\begin{subfigure}{\columnwidth} 
\centering
\caption{The actor's goal is to deliver two packages to the delivery area:}
\includegraphics[width=0.5\columnwidth]{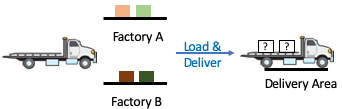}
\label{fig:ex1}
\end{subfigure} 
\begin{subfigure}{\columnwidth}
\centering
\caption{Plan-1 - Actor delivers 1 package from factory A and 1 from B:}
\includegraphics[width=0.8\columnwidth]{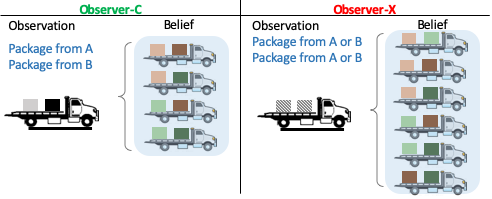}
\label{fig:ex2}
\end{subfigure}
\begin{subfigure}{\columnwidth}
\centering
\caption{Plan-2 - Actor delivers both the packages from factory A, helping observer-C and thwarting observer-X:}
\includegraphics[width=0.8\columnwidth]{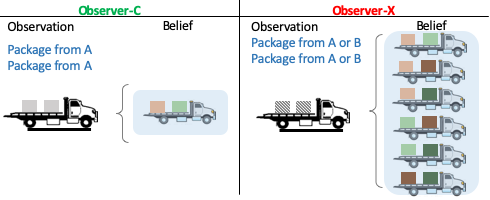}
\label{fig:ex3}
\end{subfigure}
\caption{Different belief updates induced by the same plan for two observers; observer-C is cooperative, observer-X is adversarial.}
\label{fig:example2}
\vspace{-8pt}
\end{figure}

\mysection{Motivating Example}
Consider the example in Figure \ref{fig:example2}, where a truck agent has to deliver two packages. As shown in Figure \ref{fig:ex1}, there are two packages at factory A and two at factory B. 
Let there be two observers who know the initial inventory at each factory, but do not know which two packages will be delivered by the agent. Each observer has different sensors (or sources of information) about the truck's activities. Observer-C (cooperative observer) has access to a database of barcodes which allows it to determine the identity of the factory that the package was loaded from (but not the identity of the individual packages), whereas observer-X (adversarial observer) does not have any additional information and cannot differentiate between any of the packages. The truck agent wants to convey its true goal to observer-C but obfuscate it to observer-X. 
When the agent loads one package each from A and B (Figure \ref{fig:ex2}), observer-C cannot distinguish the packages. However, when it loads both the packages from A (Figure \ref{fig:ex3}), observer-C can infer the identity of the loaded packages, while observer-X still does not know the identity of the packages. Therefore, plan-2 makes the true goal legible to observer-C while obfuscating the true goal with all the possible goals for observer-X. 

\vspace{-10pt}

\section{\textsc{mo-copp}}

\textsc{mo-copp} setting involves an actor ($\mathbf{A}$) and two observers, where one is adversarial observer ($\mathbf{X}$) while the other is cooperative ($\mathbf{C}$). We assume that the actor has full observability of its own activities and knows the sensor models used by the observers, and that the observers have different sensor models.\footnote{If the actor does not have access to the sensor models of the observers or both the observers have the same sensor model, then there is of course no question of balancing obfuscation and legibility.}  When the actor takes an action and reaches a new state, each observer receives an observation. After obtaining the observations, each observer updates its belief. The actor leverages the known limits in the observers' sensors to control the observability of multiple observers in the environment simultaneously. 
Given a set of candidate goals, the objective of the actor is to convey information about its goal to the cooperative observer and to hide it from the adversarial observer. 
This involves increasing the number of candidate goals possible in the adversary's belief, while decreasing the number of candidate goals possible in cooperative observer's belief.

\begin{dfn}
A \textbf{mixed-observer controlled observability planning problem} is a tuple, $\textsc{mo-copp} = \langle \Lambda, \mathcal{P}, \mathcal{G}, \{\Omega_i\}_{i \in \Lambda} , \{\mathcal{O}_i\}_{i \in \Lambda},$ $\{\mathcal{B}^i_0\}_{i \in \{\mathbf{X}, \mathbf{C}\}} \rangle $, where,
\begin{itemize}
\item $\Lambda = \{ \mathbf{A},\mathbf{C}, \mathbf{X}\}$ is the set of agents. 
\item $\mathcal{P} = \langle  \mathcal{F}, \mathcal{O}p, \mathcal{I}, G_A \rangle $ is $\mathbf{A}$'s task captured as a classical planning problem \cite{geffner2013concise}, where $\mathcal{F}$ is the set of fluents, $\mathcal{O}p$ is the set of actions, $\mathcal{I}$ is the initial state and goal $G_A$ is a subset of fluents. 
Also, for $a \in \mathcal{O}p$, $pre(a)$, $add(a)$, $delete(a)$ are each a subset of fluents representing precondition, add effects and delete effects of $a$.
\item $\mathcal{G} = \{G_1, G_2, \ldots, G_{n-1}, G_A\}$ is the set of candidate goals, where $G_A$ is the true goal of $\mathbf{A}$, which is not known to both $\mathbf{C}$ and $\mathbf{X}$.
\item $\Omega_i$ is the set of observation symbols for agent $i$, which are emitted when $\mathbf{A}$ takes an action and reaches a new state. Further, $\Omega_A = \{o^{\mathbf{A}}_{a,s}|  a \in \mathcal{O}p, s \in \mathcal{S}\}$.
\item $\mathcal{O}_i: \mathcal{O}p \times \mathcal{S} \rightarrow \Omega_i$ is agent $i$'s deterministic sensor model. $\mathcal{S}$ is the set of states, where each state is instantiation of all fluents. Further, $\mathcal{O}_\mathbf{A}$ 
maps each action-state pair to a unique observation, $\forall~ a, a' \in \mathcal{O}p,~ s, s' \in \mathcal{S}, a \neq a' \land s \neq s': \mathcal{O}_\mathbf{A}(a, s) \neq \mathcal{O}_\mathbf{A}(a', s')$, while $\mathcal{O}_\mathbf{X}$ and $\mathcal{O}_\mathbf{C}$ are noisy sensor models that map multiple action-state pairs to the same observation symbol.
\item $\mathcal{B}^i_0$ is the initial belief of an observer, $i \in \{\mathbf{X}, \mathbf{C}\}$. The initial belief is a set of states inclusive of $\mathcal{I}$.
\label{dfn:mo-copp-def}
\end{itemize} 
\end{dfn}

From the above definition, we see that, while $\mathbf{X}$ and $\mathbf{C}$ can have arbitrary sensor models with partial observability, $\mathbf{A}$ has full observability, due to one-to-one mapping of the sensor model. The observers are aware of the planning domain of the actor and of the candidate goals, except they do not know which candidate goal is the actor's true goal, $G_{\mathbf{A}}$. 

Every time the actor acts, each $i \in \Lambda$ receives an observation consistent with its sensor model. The sensor model of an observer $i \in \{\mathbf{X}, \mathbf{C}\}$  supports many-to-one mapping of $\langle a,s \rangle$ pairs to observation symbols, i.e., $\exists a, a' \in \mathcal{O}p, s, s' \in \mathcal{S}, a \neq a' \land s \neq s': \mathcal{O}_i(a, s) = \mathcal{O}_i(a', s')$.\footnote{Our formulation supports sensor models that yield no observations. This can be modeled by using a dummy observation symbol that essentially maps to all the possible $\langle a,s \rangle$ pairs.}
For an agent $i$, the inverse of sensor model gives the set of $\langle a,s \rangle$ pairs consistent with an observation symbol $o^i \in \Omega_i$, i.e., $\mathcal{O}_i^{-1}(o^i) = \{ \langle a, s \rangle | \forall a \in \mathcal{O}p, s \in \mathcal{S}, \mathcal{O}_i(a, s) = o^i\}$.

Each observer $i \in \{\mathbf{X}, \mathbf{C}\}$ maintains its own belief, which is a set of states. $\Gamma(\cdot)$ is a transition function, such that, $\Gamma(s, a) = \bot$ if $s \not\models pre(a)$; else $\Gamma(s, a) = s \cup add(a) \setminus delete(a)$. Now we can define the belief update: (1) at time step $t = 0$, the initial belief of observer $i$ is given by  $\mathcal{B}^i_0$, (2) at time step $t \in \{1, \ldots, \mathcal{T}\}$, $\mathcal{B}^i_{t} = \{ \hat{s}~|~ \exists \hat{a},\bar{s}~~ \Gamma(\bar{s},  \hat{a}) =  \hat{s}; \bar{s} \in \mathcal{B}^i_{t-1}; \mathcal{O}_i(\hat{a}, \hat{s}) = o^i_t\}$, where $\mathcal{T}$ is the time horizon.   
That is, the belief is updated using the previous belief and the observation received. 
A plan $\pi$ (or sequence of actions) is associated with a sequence of observations, $ObsSeq_i(\pi) = \{o^i_1, \ldots, o^i_\mathcal{T}\}$ for each observer $i$. The belief of $\mathbf{A}$ is a single state due to full observability.

\mysection{\textsc{mo-copp} Solutions} A plan $\pi$ solves a $\textsc{mo-copp} = \langle \Lambda, \mathcal{P},$ $\mathcal{G}, \{\Omega_i\}_{i \in \Lambda} , \{\mathcal{O}_i\}_{i \in \Lambda}, \{\mathcal{B}^i_0\}_{i \in \{\mathbf{X}, \mathbf{C}\}} \rangle $, iff: $~\Gamma(\mathcal{I}, \pi) \models G_A$. In other words, any solution to $\mathcal{P}$ is a solution to \textsc{mo-copp}. We measure the quality of a \textsc{mo-copp} solution in terms of the extent to which the plan is legible to $\mathbf{C}$ and obfuscating to $\mathbf{X}$. Given a solution plan, $\pi$, let $\mathcal{G}^\pi_\mathbf{X} = \{G \in \mathcal{G}: \exists s \in \mathcal{B}^\mathbf{X}_\mathcal{T},~ s \models G \}$, denote the goals present in $\mathbf{X}$'s final belief and $\mathcal{G}^\pi_\mathbf{C} = \{G \in \mathcal{G}: \exists s \in \mathcal{B}^\mathbf{C}_\mathcal{T},~ s \models G \}$, denote the goals present in $\mathbf{C}$'s final belief. Also, we can define the inverse of its observation sequence for observer $i$, as follows: $\forall i \in \{\mathbf{X}, \mathbf{C}\}, ObsSeq^{-1}_i(\pi) = \{ \hat{\pi} | \exists s^i_0 \in \mathcal{B}^i_0, G \in \mathcal{G}^\pi_i, \Gamma(s^i_0, \hat{\pi}) \models G, ObsSeq_i(\pi) =  ObsSeq_i(\hat{\pi}) \}$.  

Here $|\mathcal{G}^\pi_\mathbf{X}|$ represents the amount of \textbf{goal obfuscation for $\mathbf{X}$}, and by increasing the number of goals in $\mathcal{G}^\pi_\mathbf{X}$, we can increase the goal obfuscation. Similarly, $|\mathcal{G}^\pi_\mathbf{C}|$ represents the amount of \textbf{goal legibility for $\mathbf{C}$}, and by decreasing the number of goals in $\mathcal{G}^\pi_\mathbf{C}$, we can increase the goal legibility. We now introduce a goal difference metric to estimate the quality of a solution plan to \textsc{mo-copp}.

\begin{dfn}
Given a solution plan $\pi$ that solves \textsc{mo-copp}, where $|\mathcal{G}| = n$, the \textbf{goal difference}, of $\pi$ is given by:
\begin{math}
GD(\pi) = \frac{|\mathcal{G}^\pi_\mathbf{X}| - |\mathcal{G}^\pi_\mathbf{C}|}{n-1}
\end{math}
where the denominator represents the difference between ideal values of $\mathcal{G}^\pi_\mathbf{X}$ and $\mathcal{G}^\pi_\mathbf{C}$.
\end{dfn}

An optimal solution to \textsc{mo-copp} maximizes the trade-off between amount of goal obfuscation and goal legibility. That is, it maximizes the difference between the number of goals in $\mathcal{G}^\pi_\mathbf{X}$ and $\mathcal{G}^\pi_\mathbf{C}$. 
Equivalently, closer the $GD(\pi)$ value to 1, better is the plan quality. A solution plan with $GD(\pi) = 1$ is an optimal plan. E.g., in Figure \ref{fig:ex3}, plan-2 is an optimal plan with $GD(plan$-$2) = \frac{6-1}{6-1} = 1 $. The denominator is essential for comparing the quality of plans across different problems, with varying number of candidate goals. 

\begin{prop}
Given a solution plan, $\pi$, to \textsc{mo-copp}, if $|\mathcal{G}^\pi_\mathbf{C}|=1$, then $G_\mathbf{A} \in \mathcal{G}^\pi_\mathbf{C}$.
\end{prop}

The above proposition states that when maximum goal legibility is achieved, only one goal $G_\mathbf{A}$ is present in $\mathbf{C}$'s final belief. This is because $\mathcal{I} \in \mathcal{B}^\mathbf{C}_0$, and while performing belief update at $t$, $\langle a_t, s_t \rangle \in \mathcal{O}_{\mathbf{C}}^{-1}(o^\mathbf{C}_t$). That is the true state is part of each belief update. Therefore, $G_\mathbf{A} \in \mathcal{G}^\pi_\mathbf{C}$.

\vspace{-8pt}
\section{Plan Computation}

We now present two solution approaches. 
In the first approach, we formulate \textsc{mo-copp} as a constraint optimization problem and provide an IP encoding to solve it in $\mathcal{T}$ steps. The IP encoding provides an optimal solution for the given horizon by maximizing the trade-off between the amount of obfuscation and legibility. In the second approach, we use a heuristic-guided forward search to achieve preset levels of goal obfuscation and legibility. The search algorithm generates solutions that satisfy a prespecified lower bound on the amount of goal obfuscation and goal legibility.

\vspace{-7pt}
\subsection{\textsc{mo-copp} as Integer Program} 


\mysection{Variables} We require the following binary variables for our encoding: $x_{a, t}$ is an indicator variable for action $a$ at time $t$, $y_{s, t}$ is an indicator variable for state $s$ at time $t$, $w^i_{o, t}$ is an indicator variable for observation $o^i$ at time $t$, $b^{i}_{s, t}$ is an indicator variable for state $s$ in belief $\mathcal{B}^i$ at time $t$, $h^{i}_{s, a, t}$ is an indicator variable for action $a$ being applicable in state $s$ in belief $\mathcal{B}^i$ at time $t$ and $g^{i}_{G, \mathcal{T}}$ is an indicator variable for a goal $G$ present in belief $\mathcal{B}^i_\mathcal{T}$.


\mysection{Objective Function} The objective function is essentially the numerator of $GD(\cdot)$ metric, i.e., 
\begin{math}
max~~ \sum_{G \in \mathcal{G}}~ g^{\mathbf{X}}_{G, \mathcal{T}} -~ \sum_{G \in \mathcal{G}}~ g^{\mathbf{C}}_{G, \mathcal{T}}
\label{eq:1}
\end{math}.
We skip the denominator of the $GD$ metric, as it is a constant and does not contribute to the optimization. 
This provides a single solution that achieves the maximum difference between the number of goals possible for the two observers. 
Note that, it would make sense to get the Pareto optimal solutions if we wanted to explore all the combinations of goals achieved for the two observers. However, that is not our objective.

\mysection{IP Constraints}

\vspace{-8pt}

\begin{scriptsize}
\begin{align}
&\forall s \in \mathcal{S}, s = \mathcal{I}:~ y_{s, 0} = 1;~ s  \neq \mathcal{I}:~ y_{s, 0} = 0;~ \sum_{G_A \in~ s} y_{s, \mathcal{T}} = 1 \label{eq:3} \\
&\forall i   \in \{\mathbf{X}, \mathbf{C}\}, s \in \mathcal{S}, s \in \mathcal{B}^{i}_0:~ b^{i}_{s, 0} = 1;~ s \not\in \mathcal{B}^{i}_0:~ b^{i}_{s, 0} = 0 \label{eq:6} \\
&\forall i  \in \{\mathbf{X}, \mathbf{C}\}, G \in \mathcal{G}, m > |\{s|~ G \in s \}|:~ m*g^i_{G, \mathcal{T}} - \sum_{G~ \in~ s} b^{i}_{s, \mathcal{T}} \geqslant 0  \label{eq:7} \\
&\forall a \in \mathcal{O}p, t \in \{ 1, \ldots, \mathcal{T} \},~ pre_a = \{s ~|~ pre(a) \in s \}: \nonumber \\
&\quad x_{a, t} \leqslant \sum_{s \in pre_a} y_{s, t-1} \label{eq:8}\\
&\forall s, s' \in \mathcal{S}, t \in \{ 1, \ldots, \mathcal{T} \}, add_{s'} = \{a | pre(a) \in s, add(a) \setminus delete(a) \in s'\}, \nonumber \\
&\quad pre_{s'} = \{s |~ pre(a) \in s \land add(a) \setminus delete(a) \in s'\}: \nonumber \\
&\quad \sum_{a \in add_{s'}} x_{a, t} + \sum_{s \in pre_{s'}} y_{s, t-1} - 2~ y_{s', t} \geqslant 0 \label{eq:9} \\
&\forall a \in \mathcal{O}p, t \in \{1, \ldots, \mathcal{T}\}, 
post_a = \{s^{\prime} ~|~ add(a) \setminus delete(a) \in s^{\prime} \}: \nonumber \\
&\quad \sum_{s \in pre_a, s^{\prime} \in post_a} y_{s, {t-1}}~ y_{s^{\prime}, t} = x_{a, t} \label{eq:10} \\
&\forall i  \in \{\mathbf{X}, \mathbf{C}\}, o \in \Omega_i, t \in \{ 1, \ldots, \mathcal{T} \}:~ w^i_{o, t} = \sum_{a, s^\prime \in O^i_{o}} x_{a, t}~ y_{s^\prime, t}~  \label{eq:11} \\
&\forall i  \in \{\mathbf{X}, \mathbf{C}\}, s \in \mathcal{S},  t \in \{ 1, \ldots, \mathcal{T} \},~a \in add_{s}, \nonumber \\
&\quad add_{s} = \{a |~pre(a) \in s \}:~ b^{i}_{s, t-1} + w^i_{o, t} - h^{i}_{s, a, t} \leqslant 1~~~  \label{eq:12} \\
&\forall i  \in \{\mathbf{X}, \mathbf{C}\}, s \in \mathcal{S}, o \in \Omega_i, t \in \{ 1, \ldots, \mathcal{T} \},  a \in add_{s}, \nonumber  \\
&\quad add_{s} = \{a |~pre(a) \in s \}: h^{i}_{s, a, t} - b^{i}_{s, t-1} \leqslant 0 \label{eq:13} \\
&\forall i  \in \{\mathbf{X}, \mathbf{C}\}, s \in \mathcal{S}, t \in \{ 1, \ldots, \mathcal{T} \}, a \in add_{s}, s' \in post_{s}~ \nonumber \\
&\quad add_{s} = \{a |~pre(a) \in s \}, post_{s} = \{s' |~ add(a) \setminus delete(a) \in s' \}:  \nonumber \\
&\quad h^{i}_{s, a, t} - b^{i}_{s', t} \leqslant 0 \label{eq:14} \\
&\forall i  \in \{\mathbf{X}, \mathbf{C}\}, s \in \mathcal{S}, o \in \Omega_i, t \in \{ 1, \ldots, \mathcal{T} \}, a \in add_{s},  \nonumber  \\
&\quad add_{s} = \{a |~pre(a) \in s \}:~ h^{i}_{s, a, t} - w^{i}_{o, t} \leqslant 0 ~~~ \label{eq:15} \\
&\forall i  \in \{\mathbf{X}, \mathbf{C}\}, s, s' \in \mathcal{S}, t \in \{ 1, \ldots, \mathcal{T} \}, \nonumber \\
&\quad add_{s^\prime} = \{a |~ pre(a) \in s \land add(a) \setminus delete(a) \in s'\}, \nonumber \\
&\quad pre_{s^\prime} = \{s |~ pre(a) \in s \land add(a) \setminus delete(a) \in s'\}:   \nonumber \\
&\quad \sum_{s \in pre_{s^\prime}, a \in add_{s^\prime}} h^{i}_{s, a, t} - b^{i}_{s^\prime, t} \geqslant 0  \label{eq:16} \\
&\forall t \in \{ 1, \ldots, \mathcal{T} \}: \sum_{a \in \mathcal{O}p} x_{a, t} \leqslant 1~ \label{eq:17} 
\end{align}
\end{scriptsize}


\vspace{-8pt}

Constraint \eqref{eq:3} sets the initial state and says that a state that satisfies the true goal should be achieved in the last time step for $\mathbf{A}$. Constraint \eqref{eq:6} sets the initial belief for both the observers. Constraint \eqref{eq:7} says that if a goal is satisfied in the final belief of an observer then the corresponding goal variable will be true. Constraint \eqref{eq:8} through \eqref{eq:10} enforce the transition function. Specifically, constraint \eqref{eq:8} validates the applicability of an action in a state, constraint \eqref{eq:9} states that for a resulting state to be true both the action and the state in which it is applied should be true, and similarly constraint \eqref{eq:10} validates an action with respect to its previous state and the resulting state. Constraint \eqref{eq:11} enforces the corresponding observation symbol for each observer depending on the $\langle a, s' \rangle$ pair. Constraints \eqref{eq:12} through \eqref{eq:16} enforce a belief update. Specifically, constraint \eqref{eq:12} states that an action is not applicable in a belief state if either the belief state or the observation is untrue. Constraint \eqref{eq:13} states that an action cannot be applied in a belief state that is untrue. Constraint \eqref{eq:14} states that an action cannot be true if the resulting belief state is untrue. Constraint \eqref{eq:15} states that an action cannot be true if the corresponding observation is untrue. Constraint \eqref{eq:16} states that a belief state is true if the sum of actions leading to it is at least 1. Constraint \eqref{eq:17} ensures only one action is possible at each step. 

\begin{prop}
The IP encoding listed above which takes time horizon $\mathcal{T}$ as input, solves a $\textsc{mo-copp} = \langle \Lambda, \mathcal{P}, \mathcal{G}, \{\Omega_i\}_{i \in \Lambda} , \{\mathcal{O}_i\}_{i \in \Lambda},$ $\{\mathcal{B}^i_0\}_{i \in \{\mathbf{X}, \mathbf{C}\}} \rangle$ such that, the following properties hold:
\begin{itemize}
\item \textbf{Soundness}: A solution to the IP will solve $\textsc{mo-copp}$.  
\item \textbf{Completeness}: If there exists a plan that solves the $\textsc{mo-copp}$ in $\mathcal{T}$ time steps, then it will be a feasible solution for the IP encoding. 
\item \textbf{Optimality}: An optimal solution to the IP encoding will be a plan that solves the $\textsc{mo-copp}$ with an optimal value of $GD$ given the time horizon, $\mathcal{T}$.
\end{itemize}
\end{prop}

Since a solution to the IP, $\pi_{IP}$, satisfies \eqref{eq:3}, it solves $\textsc{mo-copp}$. If there exists a plan $\pi_\mathcal{T}$ of $\mathcal{T}$ time steps that solves \textsc{mo-copp},  $\Gamma(\mathcal{I}, \pi_\mathcal{T}) \models G_A$, then $\pi_\mathcal{T}$ will satisfy the constraints \eqref{eq:3}, \eqref{eq:8}-\eqref{eq:10} and \eqref{eq:17}. The IP encoding uses the numerator of $GD$ metric as its objective function. An optimal solution to the IP encoding, $\pi^*_{IP}$, is an optimal plan to $\textsc{mo-copp}$ given plans of $\mathcal{T}$ time steps. 

\mysection{Goal Constraints} The objective function trades off goal obfuscation with goal legibility for the observers. However, the actor can ensure a predefined level of goal obfuscation (say obfuscate with \emph{at least} $k$ candidate goals) by \begin{math}
\sum_{G \in \mathcal{G}} g^{\mathbf{X}}_{G, \mathcal{T}} \geqslant k, ~~~s.t.~ 1 \leqslant k \leqslant |\mathcal{G}| 
\end{math} 
and goal legibility (say legible with \emph{at most} $j$ goals) by 
\begin{math}  
\sum_{G \in \mathcal{G}} g^{\mathbf{C}}_{G, \mathcal{T}} \leqslant j, ~~~s.t.~ 1 \leqslant j \leqslant |\mathcal{G}| 
\end{math}. 
These constraints allow the actor to filter out solutions that do not satisfy minimum bound for goal obfuscation and goal legibility. The actor can improve the robustness of the plans generated by using these constraints.

\vspace{-6pt}

\subsection{Search Algorithm}

In this section, we show that it is possible to leverage search techniques that address goal obfuscation and goal legibility in isolation to solve \textsc{mo-copp}. We adapt \citet{implicitHRC2018} approach to address goal obfuscation and goal legibility simultaneously to two different observers. We specify bounds on the amount of goal obfuscation and goal legibility desired, similar to the ones seen in the IP: obfuscate with at least $k$ goals, make it legible with at most $j$ goals. These bounds, $\Phi = \langle \Phi_\mathbf{X}, \Phi_\mathbf{C}\rangle$, are given as input to the search algorithm. 

\begin{algorithm}
    \caption{Heuristic-Guided Search}
    \label{alg:algorithm1}
    \begin{algorithmic}[1]
    \begin{scriptsize}
    \STATE Initialize $open$, $closed$ and $temp$ lists; $\Delta = 1$ 
    \STATE $\langle b^X_{\Delta}, b^C_{\Delta} \rangle \gets$ approx$(\mathcal{I}, \mathcal{B}^X_0, \mathcal{B}^C_0)$ 
    \STATE $open.push(\mathcal{I}, \langle b^X_{\Delta}, b^C_{\Delta} \rangle, \langle \mathcal{B}^X_0, \mathcal{B}^C_0 \rangle, priority = 0)$
    \WHILE{$\Delta \leqslant |\mathcal{S}|$}
        \WHILE{$open \neq \emptyset$}
        \STATE $s, \langle b^X_{\Delta}, b^C_{\Delta} \rangle, \langle \mathcal{B}^X, \mathcal{B}^C \rangle, h_{node} \gets open.pop() $
            \IF{$ |b^X_{\Delta}| > \Delta$ or $|b^C_{\Delta}| > \Delta$}
                \STATE $temp.push(s, \langle b^X_{\Delta}, b^C_{\Delta} \rangle, \langle \mathcal{B}^X, \mathcal{B}^C \rangle,  h_{node})$
                \STATE continue
            \ENDIF
            
            \STATE add $\langle b^X_{\Delta}, b^C_{\Delta} \rangle$ to $closed$
            \IF{$s \models G_A$ and $\mathcal{B}^X \models \Phi_X$ and $\mathcal{B}^C \models \Phi_C$}
            \STATE $\textbf{return}~\pi, ObsSeq_{X}(\pi), ObsSeq_{C}(\pi)$
            \ENDIF
            
            \FOR{$s^{\prime} \in $successors$(s)$}
            \STATE $o^X \gets \mathcal{O}_X(a, s^{\prime})$; $o^C \gets \mathcal{O}_C(a, s^{\prime})$ 
            \STATE $\widehat{\mathcal{B}^X}=$ Update($\mathcal{B}^X, o^X);  \widehat{\mathcal{B}^C} =$ Update($\mathcal{B}^C, o^C)$
            \STATE $\langle \widehat{b^X_{\Delta}}, \widehat{b^C_{\Delta}} \rangle \gets$ approx$(s', \widehat{\mathcal{B}^X}, \widehat{\mathcal{B}^C)}$ 
            \STATE $h_{node} \gets h_{G_A}(s') + h_{\mathcal{G}_{k-1}}(\widehat{\mathcal{B}^X}) - h_{\mathcal{G}_{\mathcal{G}-j}}(\widehat{\mathcal{B}^C})$ 
            \STATE add new node to $open$ if $\langle \widehat{b^X_{\Delta}}, \widehat{b^C_{\Delta}} \rangle$ not in $closed$\\
            \ENDFOR
            
        \ENDWHILE
        \STATE increment $\Delta$; copy items from $temp$ to $open$; empty $temp$ \\
    \ENDWHILE
    \end{scriptsize}
\end{algorithmic}
\end{algorithm}

Each search node maintains the associated beliefs for both observers. The $approx$ function generates an approximate belief, $b^i_\Delta$, of size $\Delta$ (i.e. cardinality of $b^i_\Delta$ is $\Delta$). $b^i_\Delta$ is always inclusive of the true state of the actor, this is because the actor can only take actions that are consistent with its true state. If all such $\Delta-$sized beliefs are explored then $b^i_\Delta$ of $\Delta+1$ size is computed, and this node gets put in the temporary list and is explored in the next outer iteration when $\Delta$ has been incremented. For each $\Delta$, all $\Delta$-sized unique combinations of belief (that include the actual state of the actor) are explored. This allows systematic and complete exploration of multiple paths to a given search node. 
The inner iteration performs heuristic guided forward search (we use greedy best first search) to find a plan while tracking at most $\Delta$ states in each $b^i_\Delta$. In the inner loop, the node expansion is guided by (1) customized heuristic function, which computes value of the node based on true goal and belief constraints given by $\Phi$ for the observers, and (2) goal test, which checks for satisfaction of true goal and satisfaction of the belief constraints given by $\Phi$. 
The algorithm stops either when a solution is found or when all the $\Delta$ iterations have been explored.



\begin{prop}
The search algorithm listed above which takes goal-constraints $\Phi$ as input, solves a $\textsc{mo-copp} = \langle \Lambda, \mathcal{P},$ $\mathcal{G}, \{\Omega_i\}_{i \in \Lambda},$ $\{\mathcal{O}_i\}_{i \in \Lambda},$ $\{\mathcal{B}^i_0\}_{i \in \{\mathbf{X}, \mathbf{C}\}} \rangle$ such that, the following properties hold:
\begin{itemize}
\item \textbf{Soundness} Any solution to the search algorithm is a plan that solves the \textsc{mo-copp}.
\item \textbf{Completeness} If there exists a plan that solves \textsc{mo-copp} given goal-constraints $\Phi$, it will be found by the search.
\end{itemize}
\end{prop}

A solution to the search algorithm solves \textsc{mo-copp}, since the goal test ensures the true state of $\mathbf{A}$ satisfies $G_A$. The search algorithm necessarily terminates in $|\mathcal{S}|$ iterations of the $\Delta$ parameter. The $\Delta$ parameter allows systematic exploration of unique $\Delta$-sized combinations of belief, starting with $\Delta = 1$ until a solution is found or the solution space is explored, $\Delta = |\mathcal{S}|$. The goal test checks for satisfaction of constraints in $\Phi$. Hence, a plan that solves \textsc{mo-copp} given $\Phi$ will be found by the search algorithm.

\mysection{Property} In both the solution approaches, we can assert a lower bound on the extent of goal obfuscation and goal legibility for a \textsc{mo-copp} solution plan. 
In IP, we can specify the aforementioned goal constraints to assert this minimum value, while in the search, the goal tests allow us to assert it. By setting $k$, $j$ to desired values, we can eliminate solutions with low $GD$ score. This affords the following guarantee:

\begin{prop}
\label{prop:min}
Let $\mathbf{X}$ and $\mathbf{C}$ be perfectly rational adversarial and cooperative observers respectively. Given a \textsc{mo-copp} = $\langle \Lambda, \mathcal{P}, \mathcal{G},$ $\{\Omega_i\}_{i \in \Lambda}, \{\mathcal{O}_i\}_{i \in \Lambda}, \{\mathcal{B}^i_0\}_{i \in \{\mathbf{X}, \mathbf{C}\}} \rangle$ with equally probably goals, $|\mathcal{G}|=n$, and goal constraints of at least $k$ goal obfuscation for $\mathbf{X}$ and at most $j$ goal legibility for $\mathbf{C}$, then a solution plan, $\pi$, 
gives the following guarantees:
\begin{enumerate}
\item $\mathbf{X}$ can infer $G_\mathbf{A}$ with probability $\leqslant 1/k$,
\item $\mathbf{C}$ can infer $G_\mathbf{A}$ with probability $\geqslant 1/j$, and
\item Goal difference metric, $GD(\pi) \geqslant \frac{k-j}{n-1}$
\end{enumerate} 
\end{prop}

Given $|\mathcal{G}^\pi_\mathbf{X}| \geqslant k$, $X$ can infer $G_\mathbf{A}$ with probability $\leqslant 1/k$. Similarly, given $|\mathcal{G}^\pi_\mathbf{C}| \leqslant j$, $C$ can infer $G_\mathbf{A}$ with probability $\geqslant 1/j$. Also, similarly, $GD(\pi) \geqslant \frac{k-j}{n-1}$. The above proposition states that, based on the observation equivalence there is no additional information revealed about the actor's true goal. Therefore, we can derive goal detection upper bound for $\mathbf{X}$ and lower bound for $\mathbf{C}$. Also this allows us to derive a lower bound on the plan quality.

\begin{figure*}[!ht]
\begin{subfigure}{0.33\textwidth}
\centering
\includegraphics[height=1.55 in]{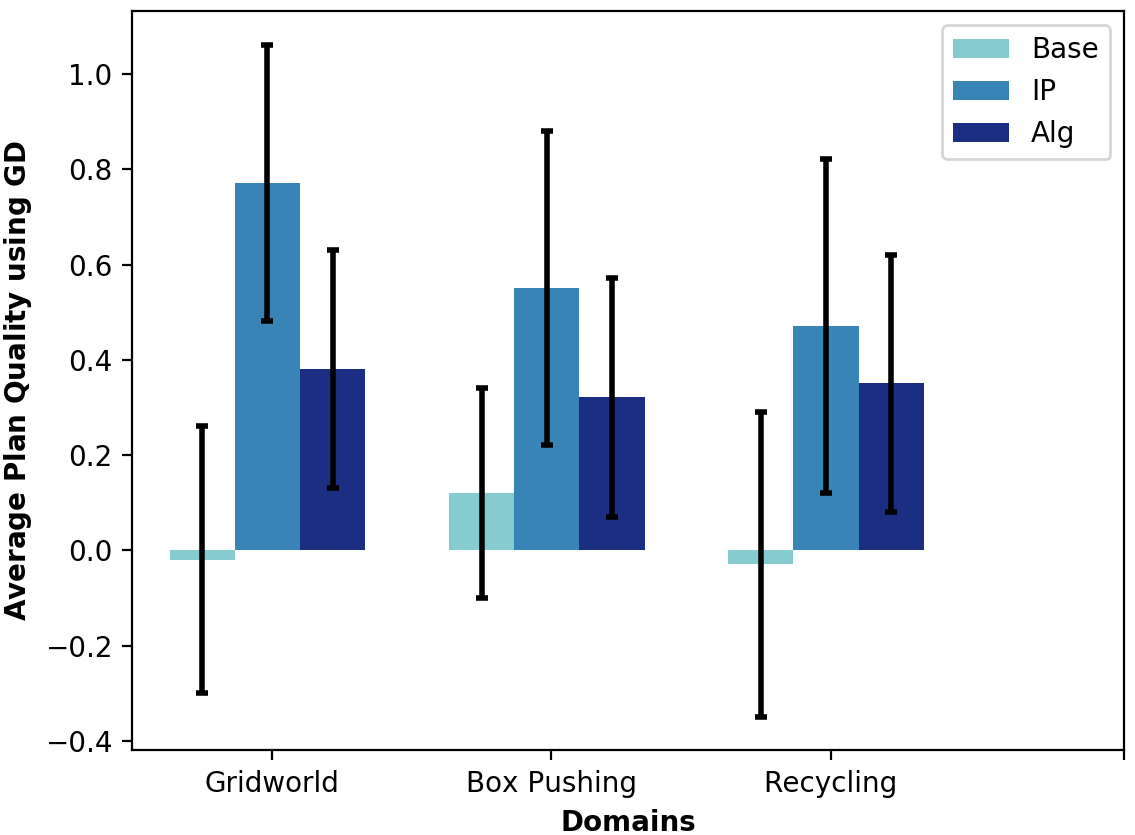}
\label{fig:graph3}
\end{subfigure}
~
\begin{subfigure}{0.33\textwidth}
\centering
\includegraphics[height=1.55 in]{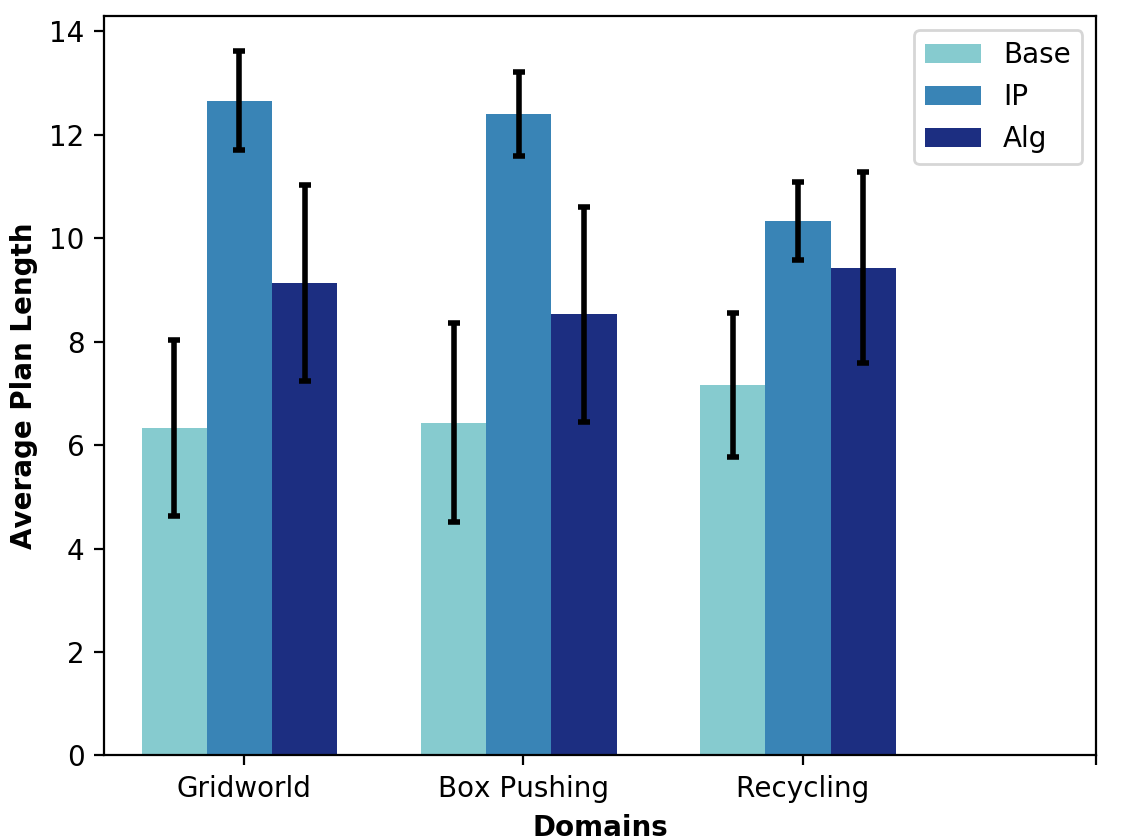}
\label{fig:graph2}
\end{subfigure}
~
\begin{subfigure}{0.33\textwidth} 
\centering
\includegraphics[height=1.55 in]{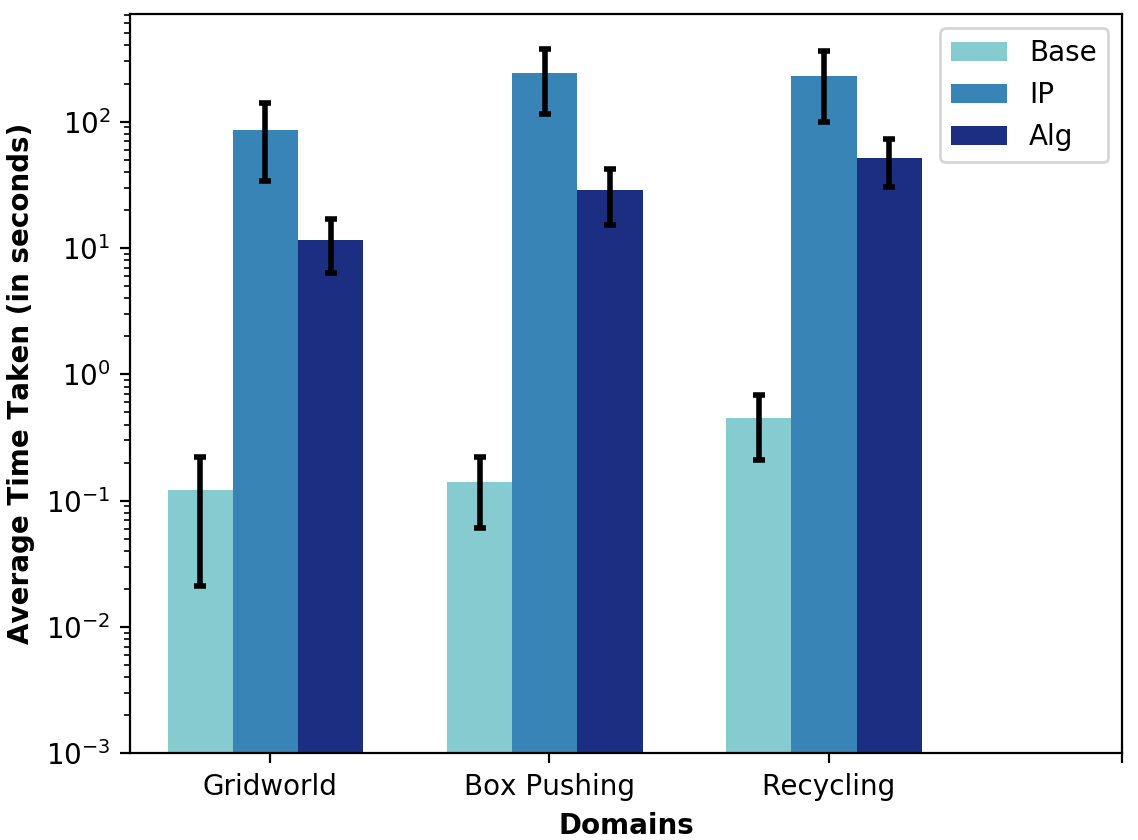}
\label{fig:graph1}
\end{subfigure} 
\caption{Comparison of average and standard deviation for goal difference (GD), plan length and run time using a baseline planner, IP planner and heuristic-guided planner over three domains.}
\label{fig:graph}
\vspace{-12pt}
\end{figure*}

\vspace{-7pt}
\section{Empirical Evaluation}

In this section, we evaluate the performance of our approaches against a baseline and discuss the relative strengths of both the approaches using 6 domains. We also compare the quality of the solutions generated by our approach against an approach that assumes either entirely adversarial or entirely cooperative observers.



\mysection{Domains} For the evaluation, we look at 6 domains: Gridworld, Box Pushing, Recycling Robot, Blocksworld, Logistics and Driverlog.
In Gridworld, the actor can move in cardinal directions. $\mathcal{O}_\mathbf{X}$, consists of two observations, \textit{vertical} for movement along N or S and \textit{horizontal} otherwise. $\mathcal{O}_\mathbf{C}$ consists of symbols, \textit{north-east} for N or E and \textit{south-west} otherwise. In Box Pushing \cite{kube1997task}, the actor's goal is to face the box and push it to the topmost row. The actions available are: \textit{move-forward, stay, turn-left, turn-right}. $\mathcal{O}_\mathbf{X}$ consists of two symbols, \textit{turn} when the agent performs turning actions and \textit{move}, otherwise. $\mathcal{O}_\mathbf{C}$ consists of \textit{move-right} when turning right or moving and \textit{leftwards} otherwise. 
In Recycling Robot \cite{sutton2018reinforcement}, the actor's objective is to collect cans and put them in  recycling bins. The actions available are: \textit{N, S, E, W, pick-up, drop, charge, stay}. $\mathcal{O}_\mathbf{X}$ consists of four symbols, \textit{horizontal} when moving E or W, \textit{vertical} when moving N or S, \textit{using-gripper} when picking or dropping and \textit{charging} otherwise. $\mathcal{O}_\mathbf{C}$ consists of \textit{north-east}, \textit{south-west}, \textit{charging-picking}, and \textit{staying-dropping}. 
For the three IPC domains, we use the lifted action names as observations for adversarial observer and lifted action names with objects as observations for cooperative observer. For example, given Blocksworld with 4 blocks \textit{a, b, c, d}, for $\mathbf{C}$, we can have \textit{stack-a-b} to represent stacking of $a$ or $b$ on any other block and similarly \textit{stack-c-d}. This is done for all the lifted actions. We produced similar observations for Logistics and Driverlog. In modeling the observations, we use the notion that in realistic scenarios, it's likely that the actor may provide the ally with tools/information to infer observations more clearly.

We implemented our IP encoding using Gurobi optimizer \cite{gurobi}. We implemented the heuristic-guided search using the STRIPS planner Pyperplan \cite{pyperplan} with hsa \cite{keyder2008heuristics} heuristic. We used hsa heuristic because it gave better performance. For baseline planner, we used greedy best first search with hsa heuristic. We ran our experiments on 3.5 GHz Intel Core i7 processor with 16 GB RAM. We used Gridworld of size 8x8, Box Pushing of size 5x5 with a single box, Recycling Robot of size 4x4 with a can and battery of 5 levels, Blocksworld with 5 blocks, and Logistics and Driverlog had goals with 4 facts each. For each domain, we generated 30 problems with random initial state goals. For the first 3 domains, we created 3 goals per problem, and for the next 3 domains, 5 goals per problem.

\mysection{Relative Strengths} 
We report the averages and standard deviation of $GD$, plan length and time taken for baseline, IP and search algorithm in Figure \ref{fig:graph}. For the baseline, we computed plans considering only $G_\mathbf{A}$. 
For the IP, we set the plan horizon to 12 for Gridworld, 12 for Box-Pushing and 10 for Recycling-Robot. If solutions were not found for that horizon, we incremented the horizon by 2. The IP did not run for the complex IPC domains, so we only display results for the first three domains. For the search, we set the minimum number of goals for $\mathbf{X}$ to 2 (at least 2 goals), and maximum number of goals for $\mathbf{C}$ to 2 (at most 2 goals). For all three algorithms, the $GD$ is calculated by counting the total number of goals present in the observers' beliefs. We tested for statistical significance of the results by performing independent measures ANOVA to reject the null hypothesis that the three algorithms are the same and that the differences are due to any randomness in the experiments (e.g., the randomly chosen goals). For Gridworld, we found that the p-value is less than 0.00001 for $GD$, as well as for plan length and run time considering the results of all 3 algorithms. This is true for Box-Pushing and for Recycling-Robot as well. All the results are significant at $p < 0.05$. 

\begin{figure}
\begin{subfigure}{0.34\columnwidth}
\centering
\includegraphics[width=1.2 in]{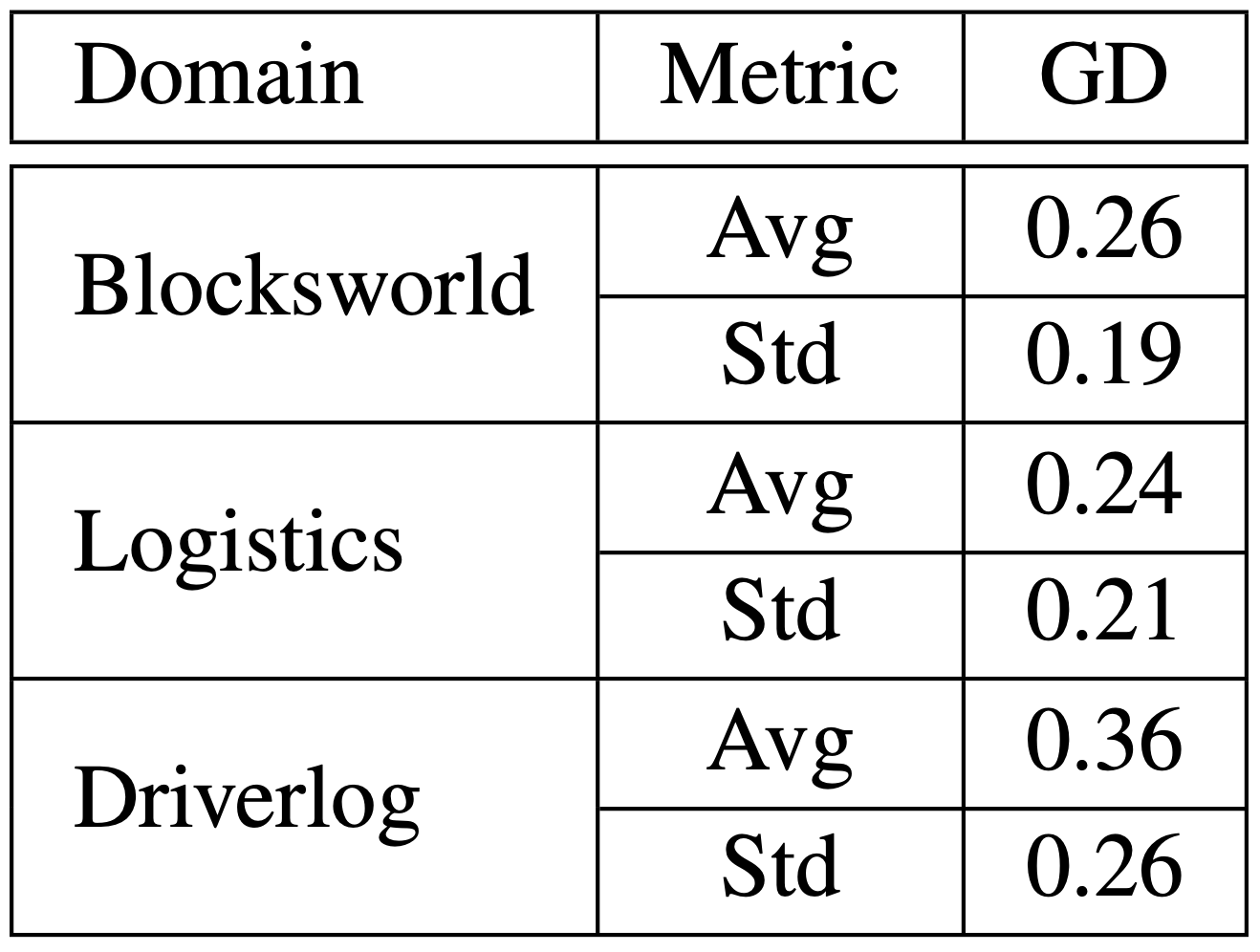}
\label{fig:graph4a}

\end{subfigure}
~
\begin{subfigure}{0.65\columnwidth}
\centering
\includegraphics[height=1.5 in]{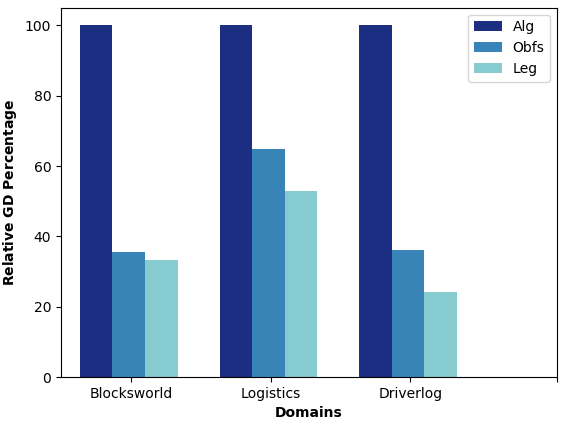}
\label{fig:graph4b}
\end{subfigure}
\caption{(a) Table shows the average and standard deviation $GD$ for IPC domains. (b) Graph shows relative $GD$ between our algorithm and approaches that achieve obfuscation/legibility in isolation.}
\label{fig:graph4}
\vspace{-8pt}
\end{figure}

The IP approach has several advantages. Firstly, it produces optimal solutions given a time horizon for the \textsc{mo-copp} problem. Secondly, it provides a lot of flexibility: it automatically chooses the best candidate goals to be added to or removed from the final beliefs of the observers. Also, if we want to specifically add or remove a particular goal from the observer's final belief, it is easy to add the necessary constraint. These advantages were evident in the results: IP has higher $GD$ for all the 3 domains. On the other hand, the search algorithm is faster and generates satisficing solutions that meet the goal constraints: the average time for search is consistently lower than IP for all the domains. Also the search solutions are shorter in length than those of the IP. The baseline although fastest (satisficing solution to a single goal) produces worst plan quality ($GD$). Additionally, the search can run more complex problems. The average and standard deviation $GD$ for the IPC domains is reported in Figure \ref{fig:graph4}.

\mysection{Comparison Against Other Approaches} 
Here, 
we report the $GD$ metric for IPC domains and show the relative performance of our algorithm against approaches that assume entirely adversarial or entirely cooperative observers using IPC domains. We used $k$-ambiguous and $j$-legible algorithms from \cite{implicitHRC2018} for achieving obfuscation and legibility in isolation as baseline. We compute the $GD$s for the baseline goal obfuscation and goal legibility by allowing minimum constraint for the other case. That is, when running goal obfuscation algorithm, the minimum constraint is to be legible with respect to at most 5 goals. Similarly, when using goal legibility algorithm, the minimum constraint is to be obfuscating with respect to at least 1 goal. We set a timeout of 20 minutes, and altogether 7 problems timed out (3 from Logistics, 4 from Driverlog) out of 90 problems. Here we set $k$ and $j$ values to 3. In Figure \ref{fig:graph4}, we report the relative $GD$ percentages for solutions that achieve goal obfuscation and goal legibility in isolation with respect to $GD$ of the solutions produced by our algorithm. The $GD$ is computed by counting the number of goals in each observer's belief. 

From Figure \ref{fig:graph4}, we can see that, our approach consistently outperforms the obfuscation and legibility algorithms with respect to the plan quality of the solutions ($GD$). This is because, as stated in Proposition \ref{prop:min}, our approach makes sure that each solution achieves a minimum amount of $GD$. In this case, the minimum is 0 (since $k$ and $j$ values are set to 3). This ensures that our search algorithm does not output solutions with $GD < 0$, which was not the case for the other two approaches, as is seen from the relative $GD$ percentages. This evaluation shows that the existing approaches that address obfuscation and legibility in isolation are not sufficient to produce good quality solutions to \textsc{mo-copp}. 


\vspace{-7pt}
\section{Related Work}

This work is related to the plan recognition literature
as the observers can use the observations to perform goal recognition. There are several prior works on goal/plan recognition \cite{ramirez2009plan,ramirez2010plan,yolanda2015fast,sohrabi2016plan,pereira2017landmark,Masters:2017}. However, the following two characteristics of our approach would not allow these goal recognition approaches to accurately rank goals: (1) our approach does not restrict the actor to optimal solutions (2) the observation equivalence in the many-to-one mapping of the observations does not provide additional information about $\langle a, s\rangle$ pairs. The aforementioned recognition systems assume 1-to-1 mapping of observations to actions or states. In our framework, given the observation sequence there exists a valid plan to each of the goals in observer's final belief. This would throw off the goal recognition systems. 
In our framework, in the adversarial case, the set of possible goals in the observer's final belief does not reveal information about the actor's true goal, whereas, in the cooperative case, this set is indeed what the actor wants to convey to the observer. 

Our framework not only accommodates both adversarial and cooperative agents but also tackles them simultaneously. However, several prior works have explored planning in adversarial environments  \cite{keren2016privacy,keren2016goal,ijcai2017-610,shekhar2018representing,ijcai2018-668,implicitHRC2018} in isolation and also in cooperative environments \cite{keren2014goal,exp-yz,explain,chakraborti2016planning,macnally2018action} in isolation. 
There are a few frameworks \cite{keren2014goal,implicitHRC2018} that are general enough to address both adversarial and cooperative settings in the same framework. However these still look at each setting in isolation in contrast to our framework. Additionally, we generalize the controlled observability planning problem \cite{implicitHRC2018} by manipulating the observations received by both adversarial and cooperative observers simultaneously. Figure \ref{fig:graph4} shows that our approaches significantly outperform \citet{implicitHRC2018} in settings with mixed adversarial and cooperative observers.

\vspace{-9pt}
\section{Conclusion}

We present the \textsc{mo-copp} formulation which is a more general framework for controlled observability planning problem. \textsc{mo-copp} can tackle both adversarial and cooperative observers simultaneously. We provide two solution approaches: (1) we formulate the problem as a constraint optimization problem and show that the \textsc{mo-copp} can be solved optimally  given the time horizon by maximizing goal obfuscation and maximizing goal legibility simultaneously to separate observers, (2) we show that it is possible to leverage frameworks that tackle obfuscation and legibility in isolation to compute satisficing solutions to \textsc{mo-copp}. We evaluate both of our approaches using 6 domains in total.


\section*{Acknowledgments}

This research is supported in part by ONR grants N00014-16-1-2892, N00014-18-1-2442, and N00014-18-1-2840, AFOSR grant FA9550-18-1-0067, NASA grant NNX17AD06G and JP Morgan AI Faculty Research grant.

\bibliographystyle{named}
\bibliography{bib}

\end{document}